\pdfoutput=1
\documentclass[11pt]{article}

\usepackage[review]{acl}

\usepackage[utf8x]{inputenc}
\usepackage{times}
\usepackage{latexsym}
\usepackage{amsthm,amsmath,amssymb}
\usepackage{mathrsfs}
\usepackage{color}
\usepackage{times}
\usepackage{booktabs}
\usepackage{amsmath}
\usepackage{float}
\usepackage{graphicx}
\usepackage{subfigure}
\usepackage{times}
\usepackage{mathrsfs}
\usepackage{color}
\usepackage{times}
\usepackage{amsmath}
\usepackage{float}
\usepackage{graphicx}
\usepackage{subfigure}
\usepackage{url}
\usepackage[noend]{algpseudocode}
\usepackage{algorithmicx}
\usepackage{algorithm}
\usepackage[T1]{fontenc}
\usepackage{microtype}

\title{Low Resource Style Transfer via Domain Adaptive Meta Learning}
\author{Xiangyang Li\textsuperscript{1,\thanks{\quad Equal contribution.}}, Xiang Long\textsuperscript{2, \footnotemark[1]},  Yu Xia\textsuperscript{1}, Sujian Li\textsuperscript{1, \thanks{\quad Corresponding author.}}\\
        \textsuperscript{1}Key Laboratory of Computational Linguistics, Peking University, MOE, China\\
        \textsuperscript{2}Beijing University of Posts and Telecommunications, Beijing, China\\
        \texttt{\{xiangyangli, yuxia, lisujian\}@pku.edu.cn}\\
        \texttt{xianglong@bupt.edu.cn}}

\begin{document}
\maketitle
\begin{abstract}
Text style transfer (TST) without parallel data has achieved some practical success. 
However, most of the existing unsupervised text style transfer methods suffer from (i) requiring massive amounts of non-parallel data to guide transferring different text styles. (ii) colossal performance degradation when fine-tuning the model in new domains. 
In this work, we propose DAML-ATM (Domain Adaptive Meta-Learning with Adversarial Transfer Model), which consists of two parts: DAML and ATM. 
DAML is a domain adaptive meta-learning approach to learn general knowledge in multiple heterogeneous source domains, capable of adapting to new unseen domains with a small amount of data.
Moreover, we propose a new unsupervised TST approach Adversarial Transfer Model (ATM), composed of a sequence-to-sequence pre-trained language model and uses adversarial style training for better content preservation and style transfer.
% can take full advantage of the sequence-to-sequence pre-trained language model to retain text content information without disentangling the content and style in the latent space.
% In this work, we propose DAML-ST5\footnote{Our code will be released after paper publication.}, which is a domain adaptive meta-learning approach and a prototype model show how to work with pre-trained generative model such as T5.
Results on multi-domain datasets demonstrate that our approach generalizes well on unseen low-resource domains, achieving state-of-the-art results against ten strong baselines.

% Compared to exist methods, our approach achieved 
%It achieves significant improvements against \uwave{six} baselines. Impressively, TSTML surpasses the in-domain performance using only \uwave{40\%} of target domain data on average.
\end{abstract}

\section{Introduction}
Text style transfer (TST) aims to change the style of the input text and keep its content unchanged, which has been applied successfully to text formalization ~\cite{jain2019unsupervised}
, text rewriting ~\cite{nikolov2018large}
, personalized dialogue generation ~\cite{niu2018polite}
and  %stylized response generation~\cite{gao2019structuring}
other stylized text generation tasks ~\cite{gao2019structuring,cao2020expertise,syed2020adapting}.

% The main reason that puzzles the development of text style transfer is the lack of parallel data. Recently, text style transfer has made great achievements in using large amount of non-parallel data~\cite{hu2017controlled,zhao2017adversarially,li2018delete}.

%注意术语的定义，让审稿人明白

%要从有的领域资源数量多，有的领域资源数量少，

%去掉非平行语料

% Most of the text style transfer models~\cite{jhamtani2017shakespearizing,wang2020formality,pryzant2020automatically} typically require a large amount of parallel sentence pairs, such as: "this soup is really hard to drink - this soup taste really delicious.", to guide the transferring of different text styles.
%The main reason that puzzles the development of text style transfer is the lack of parallel data.
%Recently, text style transfer has made great achievements in using large amount of non-parallel data~\cite{hu2017controlled,zhao2017adversarially,li2018delete}.
Text style transfer has been explored as a
sequence-to-sequence learning task using parallel datasets~\cite{jhamtani2017shakespearizing,wang2020formality,pryzant2020automatically}. However, parallel datasets are difficult to obtain due to expensive manual annotation. The recent surge of deep generative methods ~\cite{hu2017controlled,zhao2017adversarially,li2018delete} has spurred progress in text style transfer without parallel data. However, these methods typically require large amounts of non-parallel data and do not perform well in low-resource domain scenarios.

%One natural solution is to  resort to massive data from other domains,
One typical method is to resort to massive data from different domains,
which has been studied as an effective solution to address the above data insufficiency issue~\cite{glorot2011domain,wang2017instance,li2021exploring}. However, directly leveraging large amounts of data from other domains for the TST task is problematic due to the differences in data distribution over different domains, as different domains usually use their domain-specific lexica~\cite{li2019domain}.
%Different domains are usually represented in domain-specific lexica~\cite{li2019domain}.
% For instance, if we use the TST model trained on high-resource movie domain (source domain) and fine-tune it on low-resource restaurant domain (target domain), we may get unreasonable sentences like "the food is dramatic", wheree sentiment word "dramatic" is typically used in movie domain. This domain adaption issue often occurs in text style transfer due to inconsistency between the source and the target domain.
For instance, fine-tuning a TST model trained on a high-resource movie-related domain to a low-resource restaurant-related domain can get us unreasonable sentences like "the food is dramatic." The sentiment word "dramatic" is suitable for commenting a movie but weird to comment on the food.
%说daml的新在哪里，和普通maml的区别

In this work, we tackle the problem of domain adaptation in the scenarios where the target domain data is scarce and misaligned with the distribution in the source domain. Recently, model-agnostic meta-learning (MAML) has received resurgence in the context of few-shot learning scenario~\cite{lin2019personalizing, gu2018metalearning,nooralahzadeh2020zero}.
Inspired by the essence of MAML~\cite{finn2017model}, we propose a new meta-learning training strategy named domain adaptive meta-learning (DAML).  
Unlike MAML, DAML adopts a domain adaptive approach to construct meta tasks that would be more suitable to learn a robust and generalized initialization for low-resource TST domain adaption.

%Motivated by this, our key idea is to leverage the abundant data available in multiple high-resource domains to learn a robust and generalized initialization that could benefit the training process of the model in a new domain using limited target data.
% meta learning能达到这个目的，

%The goal of meta learning is to learn new tasks on the basis of known tasks with fewer steps and data. One way to achieve meta learning is to learn an optimal initialization that can accurately and quickly adapt to a new task with little data ~\cite{snell2017prototypical}. Meta learning has been applied in various situations to address low resource situation, such as personalized dialogue~\cite{lin2019personalizing}, machine translation ~\cite{gu2018metalearning}, named entity recognition~\cite{li2020metaner} and so on. To the best of our knowledge, meta learning has seldom been studied in text style transfer. 

%加入对抗训练
%把daml，在这个框架下我们设计了一个新的模型
%With the DAML strategy, we design a TST model for each domain. 
%Usually, if a TST model tries to decouple style information from the semantics of a text, it tends to produce content loss during style transfer ~\cite{hu2017toward,dai2019style,carlson2018evaluating}.
To well preserve content and transfer style, one typical strategy of a TST model is to decouple style information from the semantics of a text, and it tends to produce content loss during style transfer ~\cite{hu2017toward,dai2019style,carlson2018evaluating}.
Here, we do not try to decouple content and style, and propose a new Adversarial style Transfer model ATM, which is composed of a sequence-to-sequence pre-trained language model combined with adversarial style training for style transfer. In this way, our model can better preserve the content information without disentangling content and style in the latent space.

%Based on the DAML framework, in order to solve the problems that it is difficult to completely strip the style information from the semantics for a sentence and content information is prone to loss during style transfer~\cite{hu2017toward,dai2019style,carlson2018evaluating}, 
%we propose a new style transfer model Style-T5 (ST5), which takes full advantage of sequence-to-sequence pre-trained language models and stylistic adversarial training. 
%Based on the above advantages, ST5 can better preserve content information without disentangle the content and style in the latent space.

Combining DAML and ATM, in this paper, we propose the method named DAML-ATM, which extends traditional meta-learning to a domain adaptive method combined with a sequence-to-sequence style transfer model.
%Combining DAML training strategy and ST5 model, in this paper, we explore the use of domain adaptive meta-learning in the TST task and  propose the approach named DAML-ST5, which extends traditional meta-learning to a domain adaptive method combined with  sequence-to-sequence style transfer model ST5. 
% where a series of meta-tasks are constructed from a large pool of source domains to simulate low-resource domain adaptation task. 
DAML contains two alternating phases. During the meta-training phase, a series of meta-tasks are constructed from a large pool of source domains for balanced absorption of general knowledge, resulting in a domain-specific temporary model. 
In the meta validation stage, the temporary model is evaluated on the meta validation set to minimize domain differences and realize meta knowledge transfer across different domains. 
In ATM, a pre-training language model based TST model is used to improve text content retention. 
Moreover, we propose a two-stage training algorithm to combine the DAML training method and ATM model better. 

In summary, the
main contributions in this paper are three-fold: $(i)$ We propose a new unsupervised TST model, which achieves SOTA performance without disentangling content and style latent representations compared to other models.
$(ii)$ We extend the traditional meta-learning strategy to the domain adaptive meta transfer method, effectively alleviating the domain adaption problem in TST. $(iii)$ We propose a two-stage training algorithm to train DAML-ATM, achieving state-of-the-art performance against multiple strong baselines.

\section{Related Work}
\subsection{Text Style Transfer}

Text style transfer based on deep learning has been extensively studied in recent years. A typical pattern is first to separate the latent space as content and style features, then adjust the style-related features and generate stylistic sentences through the decoder. 
~\cite{hu2017controlled,fu2017style,li2019domain}assume that appropriate style regularization can achieve the separation.  
Style regularization may be implemented as an adversarial discriminator or style classifier in an automatic encoding process. 
However, these style transfer paradigms use large amounts of annotation data to train models for specific tasks. If we already have a model for a similar task, it is unreasonable to need many data still to train the model from scratch.

On the other hand, some of the previous work learned to do TST without manipulating the style of the generated sentence based on this learned latent space. 
~\cite{dai2019style}use the transformer architecture language model to introduce attention mechanism, but they do not make full use of the prior knowledge of sequence to sequence pre-trained language model, such as Bart~\cite{lewis2019bart} and T5~\cite{raffel2019exploring}, which have made significant progress in text generation tasks. 
In this paper, we proposed the DAML training method to solve the domain shift problem in TST and proposed a new TST model architecture named ATM, which makes no assumption about the latent representation of source sentence and takes the proven sequence-to-sequence pre-trained language model.

\subsection{Domain adaptation}
Domain adaptation has been studied in various natural language processing tasks~\cite{glorot2011domain,qian2019domain,wang2017instance,li2021knowledge}. 
However, there is no recent work about domain adaptation for a TST, except DAST~\cite{li2019domain}. 
DAST is a semi-supervised learning method that adapts domain vectors to adapt models learned from multiple source domains to a new target domain via domain discriminator. 
Different From DAST, we propose to combine meta-learning and adversarial networks to achieve similar domain adaption ability, and our model exceeds the performance of DAST without domain discriminator. Although there are some methods perform well in few shot data transfer~\cite{riley2021textsettr,krishna2021fewshot}, these methods discuss completely new text style transfer, while we focus on the domain adaptation issue.

\subsection{Model-Agnostic Meta-Learning}
% The purpose of meta-learning~\cite{schmidhuber1995learning,thrun2012learning} is to learn a general model that can quickly adapt to new tasks with a few training samples without the need to retrain from scratch. The latest meta-learning methods mainly focus on few-shot
% learning, which can be roughly divided into metric-based methods~\cite{snell2017prototypical}, memory-based methods~\cite{santoro2016meta}, and optimization-based methods~\cite{ravi2016optimization}. Some studies have applied~\cite{lee2019meta,ren2018meta,hsu2018unsupervised} apply meta-learning strategies to image classification in few-shot learning. 
Model-agnostic meta-learning (MAML)~\cite{finn2017model} provides a general method to adapt to parameters in different domains. 
MAML solves few-shot learning problems by learning a good parameter initialization. 
During testing, such initialization can be fine-tuned through a few gradient steps, using a limited number of training examples in the target domain.
% Different from traditional training methods, MAML forms a set of training tasks $T=	\left\{T_{1}, T_{2},...\right\}$, where each task consists of a training set and a validation set, and two gradient descent is used to update the model parameters.
Although there have been some researches~\cite{qian2019domain,li2020metaner,wu2020enhanced} on MAML in natural language processing, it is still scarce compared to computer vision. 
Unlike the above research on classification under few-shot learning, our research focuses on text style transfer based on text generation. 
In this paper, we seek a new meta-learning strategy combined with adversarial networks, which is more suitable for encouraging robust domain representation. 
As far as we know, we are the first to adopt meta-learning in the domain adaptation problem of text style transfer tasks.

\section{Methodology}
In this section, we first define the problem of domain adaptive learning for TST. Then we describe our approach, DAML-ATM, in detail.

\iffalse
\begin{figure}[htbp]
\centering
\includegraphics[width=7.5cm]{figure/M.png}
\caption{Detail of our base model.}
\end{figure}
\fi
\iffalse
Suppose we have two sets of style-labelled sentences
$S=\{(x_{i}^{'}, l_{i}^{'})\}_{i=1}^{N^'}$, $T=\{(x_i, l_i)\}^N_{i=1}$ in
the source domain $S$ and the target domain $T$ , respectively.
$x_i^'$ denotes the $i^{th}$ source sentence. $l_i^'$ denotes the corresponding style label, which belongs to a source style label set: ${l_i^'}
\in l^S$ (e.g.,
positive/negative). $l_i^'$
can be available or unknown. Likewise, pair $(x_i, l_i)$ represents the sentence and style label in the target domain, where ${l_i} \in {l^T}$ .

We consider domain adaptation in the following setting: the source styles are available and are
the same as the target styles, i.e., $l^T = l^S$, 
In both scenarios, we assume that the target domain
$T$ only has limited non-parallel data. With
the help of source domain data $S$, the goal is to
transfer $(x_i$$; l_i)$ to $(\tilde{x}_i$$;\tilde{l}_i)$ in the target domain.
The transferred sentence $\tilde{x}_i$ should simultaneously hold: (i) the main content with $x_i$, (ii) a different style $\tilde{l}_i$ from $l_i$, and (iii) domain-specific characteristics
of the target data distribution T .
\fi

\subsection{Task Definition} 
Let $D_{S}=\{D_1,...,D_N\}$ be $N$ source domains in the training phase, where $D_n (1 \leq n \leq N )$  is the $n$-th source domain containing style-labelled non-parallel data $D_n=\{(X_i, l_i)\}_{i=1}^{L_n}$, where $L_n$ is the total number of sentences, $X_i$ denotes the $i^{th}$ source sentence, and $l_i$ denotes the corresponding style label, which belongs to a source style label set: ${l_i}
\in L_{S}$ (e.g., positive/negative). Likewise, there are $K$ target domains $D_{T}=\{D_1,...,D_K\}$ which are unseen in $D_{S}$. Our task is to transfer a sentence $X_i$ with style $l_i$ in the target domain to another sentence $Y_i^{'}$ sharing the same content  while having a different %opposite 
style $\tilde{l}_i$ from $l_i$ and
domain-specific characteristics of the target domain.

% We apply the same setting of domain adaptation for TST as~\cite{li2019domain}: the source styles are available and are the same as the target styles. We assume that the source and  target domains only have limited non-parallel data. With the help of abundant source domain data, our objective is to transfer a sentence $x_i$ with style $l_i$ in the target domain to another sentence $x_i^'$ sharing the same content  while having a different %opposite 
% style $\tilde{l}_i$ from $l_i$ and
% domain-specific characteristics of the target domain $D_t \in DT$.

%提出两个阶段的算法

We propose a two-stage algorithm for domain adaptation in TST: pre-training learning strategy and domain adaptive meta-learning strategy. 
In pre-training learning, our objective is to make the model more able to preserve content information and distinguish between different text styles. 
In domain adaptive meta-learning, our objective is to learn a meta-knowledge learner for the sequence-to-sequence model by leveraging sufficient source
data $D_{s}$. 
Given a new unseen domain from $D_{new}$, the new learning task of TST can be solved by fine-tuning the learned sequence-to-sequence model (domain-invariant parameters) with only a small number of training samples. 
% In summary, the meta-knowledge learner of the sequence-to-sequence model should aggregate the knowledge learned from multiple domains in $D_{s}$, which results in a more robust, generic and transferable representation that can be widely used to achieve the optimum performance in $D_{new}$ with as little as possible.

\begin{figure}[htbp]
\centering
\includegraphics[scale=0.275]{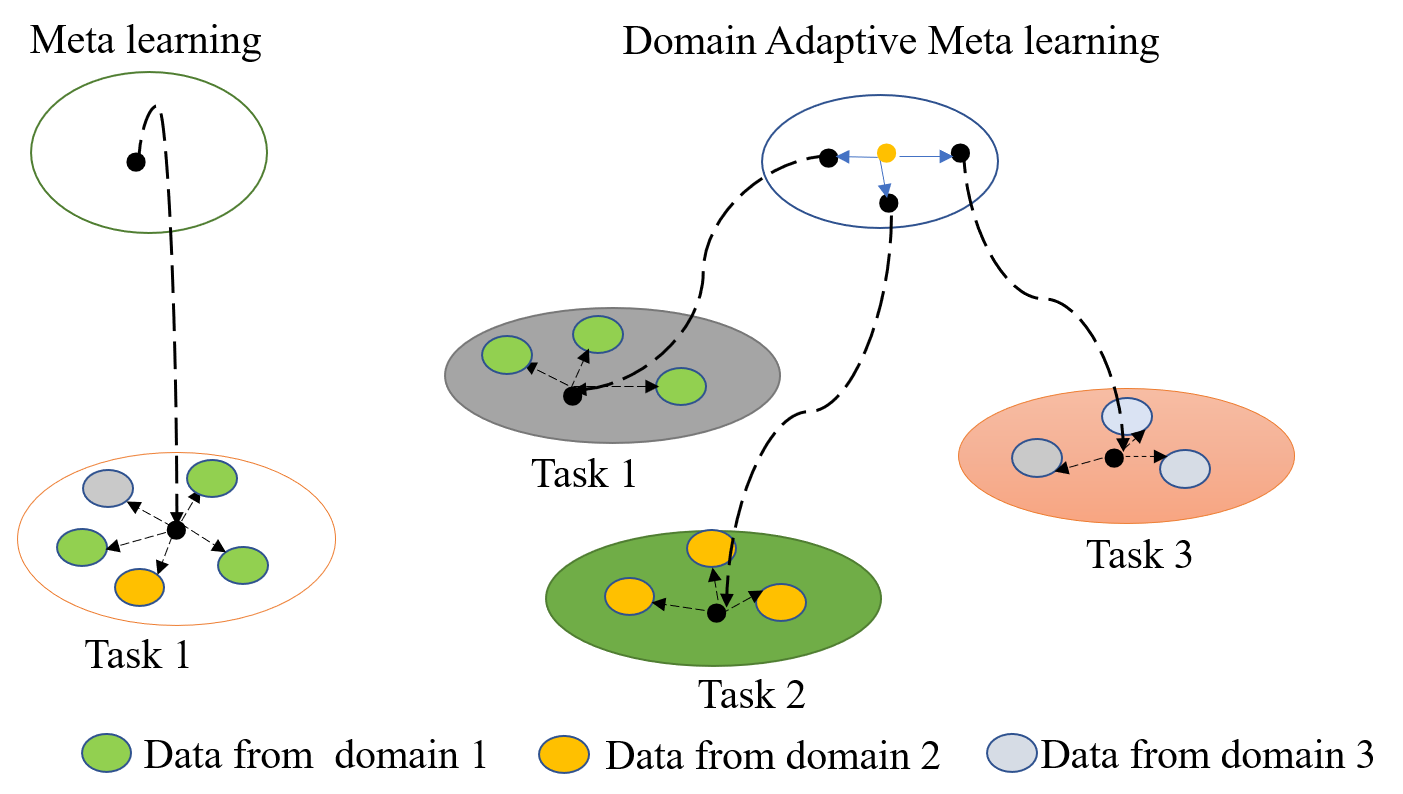}
\caption{Comparison of meta-learning and domain adaptive meta transfer learning (DAML). In DAML, each meta task contains $n$ sentences from the same domain. In MAML, the data in each meta task come from different domains.}
\label{figure1}
\end{figure}

\subsection{DAML-ATM Approach}

\subsubsection{Overview of DAML} 

Model-agnostic meta-learning can utilize a few training samples to train a model with good generalization ability. 
However, since it is based on the assumption that the meta tasks are from the same distribution (Figure~\ref{figure1}, left), simply feeding all the sources data into it might get sub-optimal results~\cite{chen2020st}. 
Therefore, we propose a modified way to construct meta tasks (Figure~\ref{figure1}, right). 

Different from MAML, for DAML, in one batch, the data in each meta task comes from the same source domain, and each meta task comes from a different domain.
In this way,  we can guarantee that DAML can learn generic representations from  different domains in a balanced way.  
During each iteration, we randomly split all source domains into a meta-training set $D_{tr}$ and a meta-validation set $D_{val}$, where $D_{S}=D_{tr}\cup D_{val}$ and $D_{tr}\cap D_{val}=\emptyset$. A meta-training task $T_i$ is sampled from $D_{tr}$ and is composed of $n$ instances from a specific domain.
Likewise, a meta-validation task $T_j$ is sampled from $D_{val}$. The validation errors on $D_{val}$ should be considered to improve the robustness of the model. 
In short,  with DAML, the parameters learned by the model in the parameter space are not biassed towards any one particular domain  with as little data as possible during model updating as shown in Figure~\ref{figure1}(right).

In the final evaluation phase, the meta-knowledge learned by the sequence-to-sequence model can be applied to new domains. 
Given a new unseen domain $D_{new} = (T_{tr}, T_{te} )$, the learned sequence-to-sequence model and the discriminator are fine-tuned on $T_{tr}$ and finally tested on $T_{te}$.

\begin{figure}[htbp]
\centering
\includegraphics[scale=0.24]{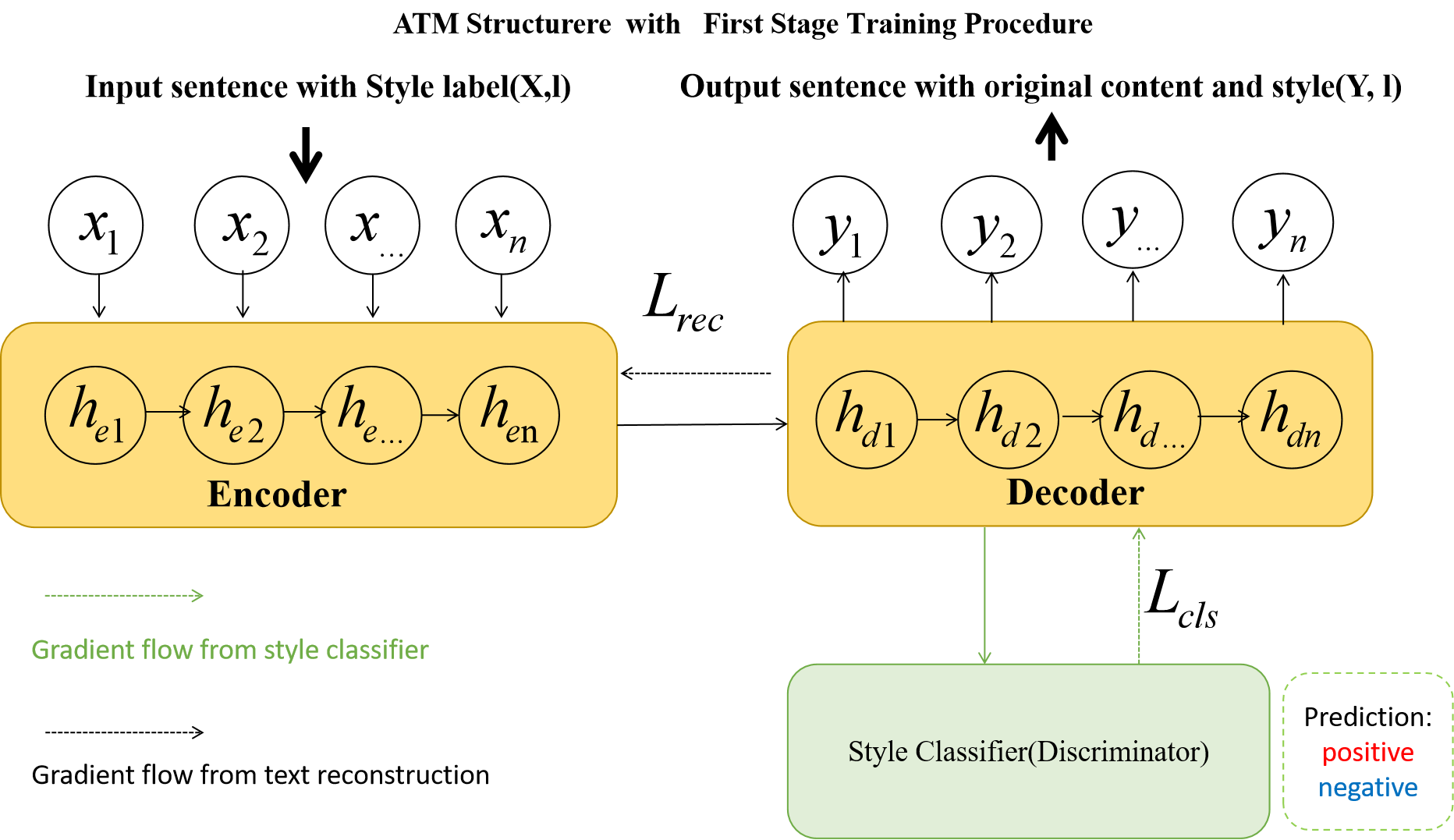}
\caption{The basic structure of our TST model, ATM, with the first stage training procedure. The green dashed line represents the loss of style classification to ensure that the style classifier can distinguish between different text styles. The black dotted line rerents text reconstruction loss to ensure the generated sentence has a similar semantic meaning as the input sentence.}
\label{figure2}
\end{figure}

% \begin{figure}[htbp]
% \centering
% \includesvg[scale=0.24]{figure/model1.svg}
% \caption{The basic structure of our TST model Style-T5 (ST5) with first stage training procedure. The green dashed line represents the loss of style classification to ensure that the style classifier can distinguish between different text styles. The black dotted line rerents  text reconstruction loss to ensure the generated sentence has a similar semantic meaning as the input sentence.}
% \label{figure1}
% \end{figure}

\subsubsection{ATM Model} 
In this section, we give a brief introduction to our proposed model: \textbf{ATM},  which combines sequence-to-sequence pre-trained model with adversarial training. (1) For the content preservation, we train the sequence-to-sequence model $\theta$ to reconstruct the original input sentence $X$  with the original style label $l$. 
(2) For the style controlling, we train a discriminator network $\gamma$ to assist the sequence-to-sequence model network in better controlling the style of the generated sentence. The structure of the model is shown in Figure~\ref{figure2}.

\textbf{S2S-model} To ease the explanation, we start with the sequence-to-sequence (S2S) model here.
% T5 is a sequence-to-sequence pre-trained language model proposed by~\cite{raffel2019exploring}, which  follows the standard transformer~\cite{vaswani2017attention} encoder-decoder architecture.
% Adopting transformer architecture~\cite{vaswani2017attention},  which is composed of an encoder Transformer and a decoder Transformer.
Explicitly, for an input sentence $X = (x_1, x_2, ..., x_n)$ of length $n$, $X\in D$, the S2S
encoder $Enc(X;\theta_{E})$ maps inputs to a sequence of continuous  hidden representations $H = (h_1, h_2, ..., h_n)$. Then, the S2S decoder $Dec(H; \theta_{D})$ estimates the conditional probability for the output
sentence $Y= (y_{1}, y_{2}, ..., y_{n})$ by auto-regressively
factorized its as:
\begin{equation}
\label{e1}
\begin{aligned}
p_{\theta }(Y|X)=\prod_{t=1}^{n}p_{\theta }(y_{t}|H, y_{1}, ..., y_{t-1})
\end{aligned}
\end{equation}
At each time step $t$, the probability of the next
token is computed by a softmax classifier:
\begin{equation}
\label{e1}
\begin{aligned}
p_{\theta }(y_{t}|H,y_{1},....,y_{t-1}))=softmax(o_{t})
\end{aligned}
\end{equation}
where $o_{t}$
is logit vector outputted by decoder network. 
The standard S2S model without discriminator makes the output sequence $Y$ the same as the input sequence $X$.
% let the output sequence $Y$ to be the same as the input sequence $X$ by teacher forcing.
% We denote an encoder-decoder
% model as $(\phi, \psi)$. The semantic representation $c_i$
% of sentence $x_i$
% is extracted by the encoder $\phi$,
% $c_i$ = $E(x_i)$. The decoder $\psi$ aims to learn a conditional distribution of $x_i$ given the semantic representation $c_i$ and style $l_i$. 
% The loss function of the sequence-to-sequence model
% minimizes the negative log-likelihood of the
% training data:

% \begin{equation}
% \label{e1}
% \begin{aligned}
% L_{rec}(\theta) = -\mathop{\mathbb{E}}\limits_{X_{i}\sim D}[log P(Y_{i}|X_{i};\theta )]
% \end{aligned}
% \end{equation}
% In sequence-to-sequence model, we let the output sequence $Y$ to be the same as the input sequence $X$.

\textbf{Discriminator Model} By teacher forcing, S2S tends to ignore the style labels and collapses to a reconstruction model, which might copy the input sentence, hence failing to transfer the style.
Therefore, to make the model learn meaningful style information, we apply a style discriminator $\gamma$ for the style regularization. 
In summary, we use a style discriminator to provide the direction (gradient) for TST to conform to the target style. 
Our discriminator is a multi-layer perceptron with a sigmoid activation function to predict style labels or guide the direction of style transfer. 
Our model training involves a pre-training learning strategy and a domain adaptive meta-learning strategy.

\subsubsection{First Stage: Pre-training Learning} 

In the first stage, we train the discriminator model to distinguish different text styles. 
In this stage, the discriminator models are equivalent to a text classifier. 
Inspired by~\cite{lewis2019bart}, we feed the hidden states from the last layer of the decoder into the classifier instead of the gumble-softmax trick~\cite{jang2017categorical} for gradient back-propagation, which is more stable and better than gumble-softmax(See Table~\ref{Table5}). 
The loss function for the discriminator is simply the cross-entropy loss of the classification problem:

% L_{cls}(\gamma) = -\mathop{\mathbb{E}}\limits_{X_{i}\sim D}[log P(l_{i}|X_{i},l_{i};\gamma )]

 \begin{equation}
\label{e1}
\begin{aligned}
\mathcal{L}_{cls}(\gamma) = -\mathop{\mathbb{E}}\limits_{X_{i}\sim D_{S}}[log P(l_{i}|X_{i},l_{i};\theta, \gamma )]
\end{aligned}
\end{equation}
For the S2S model, we pre-train the S2S model to allow the generation model to learn to copy an input sentence $X$ using teacher forcing. The loss function of the sequence-to-sequence model minimizes the negative log-likelihood of the training data:

\begin{equation}
\label{e1}
\begin{aligned}
\mathcal{L}_{rec}(\theta) = -\mathop{\mathbb{E}}\limits_{X_{i}\sim D_{S}}[log P(Y_{i}|X_{i};\theta )]
\end{aligned}
\end{equation}

In summary, we train the sequence model and the style classification model separately on the source domain to learn content preservation and style discrimination in the first stage. 
The first stage training procedure of the ATM is summarized in Algorithm~\ref{alg1}.
% \begin{figure}[htbp]
% \centering
% \includegraphics[scale=0.3]{figure/style_T5_2.png}
% \caption{The basic structure of our model style -T5. The green arrows represent the first stage and the blue arrows represent the second stage. In the first stage, the model does the text reconstruction and classification tasks. In the second stage, we use DAML method to control specific text style generation.}
% \label{figure1}
% \end{figure}

\begin{algorithm}[t]
\footnotesize
\caption{ATM Pre-traing Learning} 
% \hspace*{0.02in}
{\bf Input:} 
sequence-to-sequence model $f_{\theta}$,discriminator $\gamma$,and a dataset $D_{i}$ with style $l_{i}$ belong to $L_{s}$\\
% \hspace*{0.02in} 
{\bf Output:} 
well-trained parameter $\theta, \gamma$
\begin{algorithmic}[1]
\State Sample a batch of m sentences ${X_{1}, X_{2}, ...X_{m}}$
from $D_{i}$.
\While{in first stage and not convergence}
    \State Use $f_{\theta}$ to generate new sentence
    \State $Y_{i}$ = $f_{\theta}(X_{i},l_{i})$ 
    \State Compute $\mathcal{L}_{cls}(\gamma)$ for $Y_{i}$  by Eq. (4) ;
    \State Compute $\mathcal{L}_{rec}(\theta)$ for $Y_{i}$  by Eq. (3) ;
    % \State Serialize a task $T_j$ from the unseen domain $\mathcal{D}_{new}$ 
    % \State Update $\theta = \theta - \beta\nabla_{\theta}\sum_{T_j} L_{T_{j}}^{tr}(\theta)$
    % \State Update $\gamma = \gamma - \beta\nabla_{\gamma}\sum_{T_j} L_{T_{j}}^{tr}(\gamma)$
    \EndWhile
% \While{in second stage and not convergence}
%     \State Use $f_{\theta}$ to generate new sentence
    % \State $y^{,}$ = $f_{\theta}(x,l)$ 
    % \State Compute $L_{style}(\theta)$ for y by Eq. (6) ;
    % \State Compute $L_{rec}(\theta)$ for y by Eq. (3) ;
    % \State Serialize a task $T_j$ from the unseen domain $\mathcal{D}_{new}$ 
    % \State Update $\theta = \theta - \beta\nabla_{\theta}\sum_{T_j} L_{T_{j}}^{tr}(\theta)$
    % \State Update $\gamma = \gamma - \beta\nabla_{\gamma}\sum_{T_j} L_{T_{j}}^{tr}(\gamma)$
    % \EndWhile
    % \State return Optimal $\theta$  for test 
\end{algorithmic}
\label{alg1}
\end{algorithm}

\begin{figure}[h]

\includegraphics[scale=0.27]{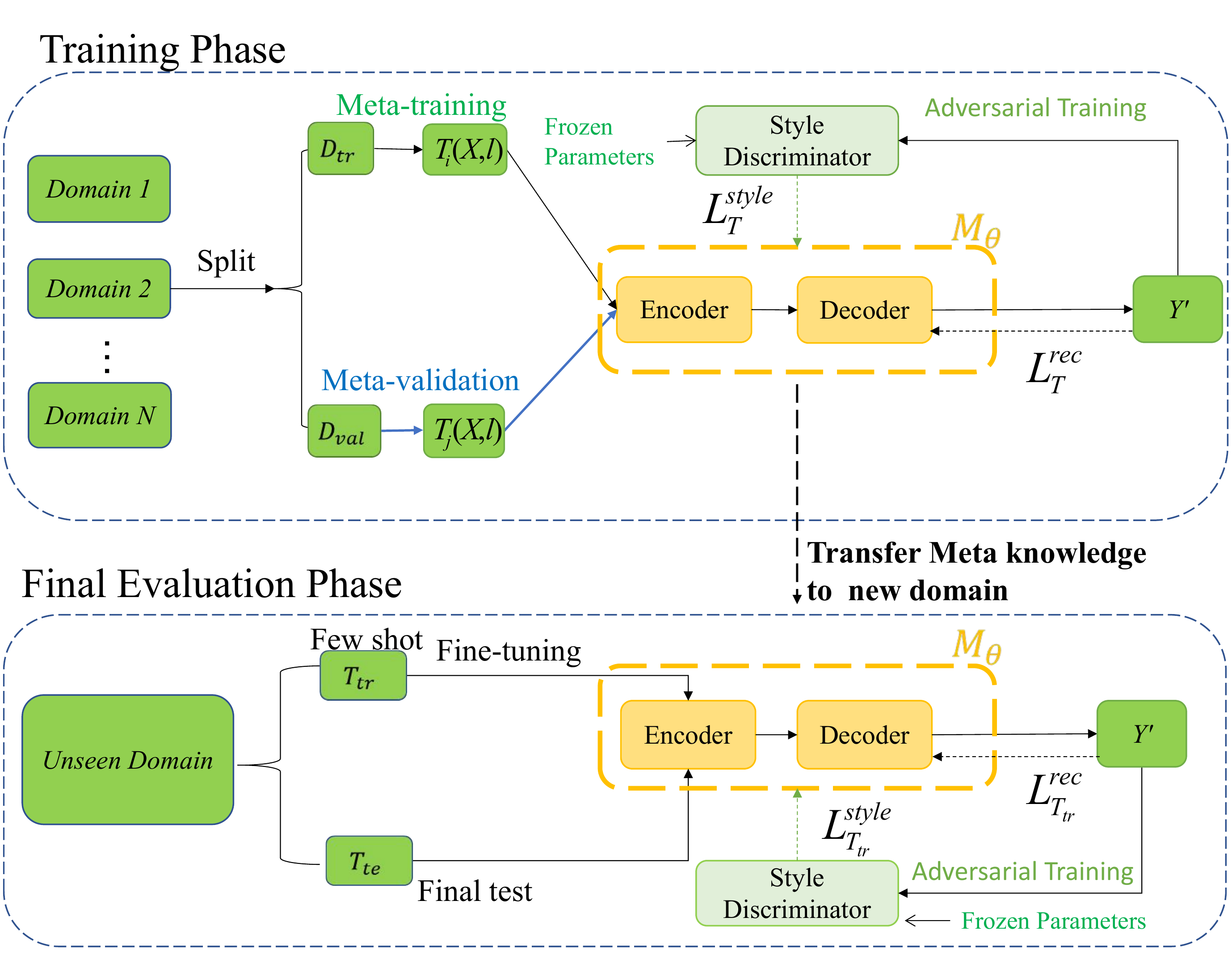}
\caption{Overview of our proposed DAML-ATM with second stage training strategy.
% DAML-ST5 explicitly simulates domain shift ($D_{tr} \rightarrow D_{val}$) during the training process. 
In the meta-training phase, a temporary model $(\theta_{old}, \theta_{new})$ is learned from $D_{tr}$. 
In the meta-validation phase, the base model is updated by gradient descent with respect to the parameters $\theta$ on $D_{val}$. 
In the final evaluation phase, the learned sequence encoder is fine-tuned on $T_{tr}$ and tested on $T_{te}$ from a unseen domain $D_{new}$. }
\label{Figure3}
\end{figure}

%对抗训练

\subsubsection{Second Stage: Domain Adaptive Meta Learning with  Adversarial Training}

After the first stage of training, the style classifier has learned how to distinguish between different text styles. 
For style controlling, we adopt a method of adversarial training to avoid disentangling the content and style in the latent space. 
The discriminator model aims to minimize the negative log-likelihood of opposite style $\tilde{l}_i$ when feeding to the sequence model sentence $X_{i}$ with the style label ${l}_i$. 
In the second stage, we freeze the parameters of the discriminator. Therefore,
style loss only works on the S2S model $\theta$, which forces the S2S model $\theta$  to generate opposite styles of sentences:

 \begin{equation}
\label{e1}
\begin{aligned}
\mathcal{L}_{style}(\theta) = -\mathop{\mathbb{E}}\limits_{X_{i}\sim D}[log P(\tilde{l}_{i}|X_{i},l_{i};\theta, \gamma )]
\end{aligned}
\end{equation}

%  \begin{equation}
% \label{e1}
% \begin{aligned}
% L_{style}(\theta) = -p_{\gamma} (c=1|f_{\theta }(x,l),l))
% \end{aligned}
% \end{equation}
% Thus, in the second stage. The  sequence-to-sequence model, guided by the style discriminator, is able to generate text in the specified style, and at the same time, the model needs to reconstruct the content information of the text. 
In the second stage, we use the DAML algorithm for domain adaptive TST, so the text reconstruction loss and the style discriminator loss are calculated over the meta-training samples in task $T_{i}$ from $D_{tr}$.
These two losses can be written as

 \begin{equation}
\label{e1}
\begin{aligned}
\mathcal{L}^{rec}_{T_{i}}(\theta) = -\mathop{\mathbb{E}}\limits_{X_{i}\sim T_{i}}[log P(Y_{i}|X_{i};\theta )] \\
\mathcal{L}^{style}_{T_{i}}(\theta) = -\mathop{\mathbb{E}}\limits_{X_{i}\sim T_{i}}[log P(\tilde{l}_{i}|X_{i},l_{i};\theta, \gamma ))
\end{aligned}
\end{equation}
% In the second stage, we freeze the parameters of the discriminator and train only the sequenc-to-sequence model. Combining the loss function we discussed above, the first stage training procedure of the Style-T5 is summarized in Algorithm 1.

% \begin{equation}
% \label{e2}
% \begin{aligned}
% L_{dis} = -\mathbb{E}_{q_{\psi}(x|c)}[log \,p_{\sigma }(\tilde{l}|x^{'})]
% \end{aligned}
% \end{equation}

We add different prefixes to the input in the second stage, which allows the S2S model to perceive different TST tasks. 
The second stage of the algorithm is called domain adaptive meta-strategy, which consists of two core phases: a meta-training phase and a meta-validation phase, as shown in Figure~\ref{Figure3}. 
% In the meta-raining and meta-validation, we adopt different learning rates ${\alpha}$, ${\beta}$ to control the parameter updates of the temporary model and the base model respectively.

\textbf{Domain Adaptive Meta-Training}

In the meta-training phase, our objective is to learn different domain-specific temporary models for each domain that are capable of learning the general knowledge of each domain.  
Inspired by feature-critic networks~\cite{li2019featurecritic}, we use a similar manner to adapt  the parameters of the domain-specific temporary model:

 \begin{equation}
\label{e1}
\begin{aligned}
\theta_i^{old} = \theta_{i-1}-\alpha\nabla \theta _{i-1}\mathcal{L}_{T_{i}}^{rec}(\theta_{i-1},\gamma _{i-1} ) \\
\theta_{i}^{new} = \theta_{i-1}^{old}-\alpha\nabla \theta_{i-1}\mathcal{L}_{T_{i}}^{style}(\theta_{i-1},\gamma_{i-1})
\end{aligned}
\end{equation}

where $i$ is the adaptation step in the inner loop, and $\alpha$ is the learning rate of the internal optimization. 
At each adaptation step, the gradients are calculated with respect to the parameters from the previous step.
For each domain of $D_{tr}$, it has different $\theta^{old}$  and $\theta^{new}$ . The base model parameters $\theta_{0}$ should not be changed in the inner loop.

\begin{algorithm}[h]
\footnotesize
\caption{The training procedure of DAML-ATM} 
% \hspace*{0.02in} 
{\bf Input:} 
${D}=\{{D}_1,...,{D}_K\}, \alpha, \beta$\\
% \hspace*{0.02in} 
{\bf Output:} 
optimal meta-learned model $\theta$
\begin{algorithmic}[1]
\State Initialize the base sequence-to-sequence model $\theta$ and discriminator model $\gamma$ by algorithm 1
\While{not converge}
   \State Randomly split ${D} =D_{tr}\cup D_{val}\ and\  D_{tr}\cap D_{val}=\emptyset$
   \State \textbf{Meta-training:}
   \For {$j$ in meta batches} \hspace*{0.2in} //Outer loop
     \State Sample a task $T_{j}$ from $D_{val}$
     \For{$i$ in adaptation steps}  \hspace*{0.2in}  //Inner loop
     \State Sample a task $T_{i}$ from $D_{tr}$
     \State Compute meta-training  rec loss $\mathcal{L}^{rec}_{T_i}$
     \State Compute meta-training style loss $\mathcal{L}^{style}_{T_j}$ 
     \State Compute adapted parameters with gradient descent for $\theta_{i-1}$
     \State $\theta_{i}^{old} = \theta_{i-1}-\alpha\nabla \theta _{i-1}\mathcal{L}_{T_{i}}^{tr}(\theta_{i-1},\gamma _{i-1} )$
\State $\theta_{i}^{new} = \theta_{i-1}^{old}-\alpha\nabla \theta_{i-1}\mathcal{L}_{T_{i}}^{style}(\theta_{i-1},\gamma_{i-1})$
        \EndFor
         \State \textbf{Meta-validation:}
        \State Compute meta-validation loss on $T_j$:  $\mathcal{L}^{val}_{T_j}$
        \EndFor
        \State \textbf{Meta-optimization}:
        \State Perform gradient step  w.r.t. $\theta$
        \State  $\theta_{0}=\theta_{0}-\beta\nabla_{\theta_{0}}\mathbb{E}_{T_j}\mathcal{L}_{T_j}^{val}(\theta_{i}^{old}, \theta_{i}^{new},\gamma)$
\EndWhile
\end{algorithmic}
\end{algorithm}

\textbf{Domain Adaptive Meta-Validation}

After meta-training phase, DAML-ATM has already learned a temporary model($\theta_{i}^{old}, \theta_{i}^{new})$ in the meta-training domains $D_{tr}$. 
The meta-validation phase tries to minimize the distribution divergence between the source domains $D_{tr}$ and simulated target domains $D_{val}$ using the learned temporary model. 
In the meta-validation phase, each temporary model is calculated on the meta-validation domain $D_{val}$ to get meta validation losses. 
 \begin{equation}
\label{e1}
\begin{aligned}
\mathcal{L}_{T_{j}}^{val}= \mathcal{L}_{T_{j}}^{rec}(\theta_{i}^{old},\gamma_{0}) + \mathcal{L}_{T_{j}}^{style}(\theta_{i}^{new},\gamma_{0})
\end{aligned}
\end{equation}
Thus, the base model $\theta$ is updated by gradient descent
 \begin{equation}
\label{e1}
\begin{aligned}
\theta_{0} = \theta_{0}-\beta\nabla \theta _{0}\mathcal{L}_{T_{j}}^{val}
\end{aligned}
\end{equation}
where $\beta$ is the meta-learning rate. 
Unlike the ordinary gradient descent process, the update mechanism of Eq. (9) involves updating one gradient by another gradient (w.r.t. the parameters of the temporary model). 
This process requires a second-order optimization partial derivative.

\subsubsection{Final Evaluation Phase of DAML-ATM} 

In the final evaluation phase, we first initialize the model with the parameters learned during the above algorithm 2. 
Then, the model takes input as a new adaptation task $T$, which consists of a small in-domain data $S_{tr}$ for fine-tuning the model and a test set $S_{te}$ for testing. The procedure is summarized in Algorithm 3. (Note that the discriminator is not needed for inference.)

\begin{algorithm}[htbp]
\footnotesize

\caption{The Final Evaluation Procedure of DAML-ATM} 
% \hspace*{0.02in}
\textbf{Input:} 
$\theta, \gamma$ learned from Algorithm 2, low resource training set $S_{tr}$ and test set $S_{te}$ of an unseen domain $D_{new}$ \\
% \hspace*{0.02in} 
{\textbf{Output:}} Performance on $S_{te}$
\begin{algorithmic}[1]
\While{ not convergence}
    \State Serialize a task $T_{tr}$ from the unseen domain ${S}_{tr}$ 
    \State Update $\theta = \theta - \beta\nabla_{\theta}\sum_{T_{tr}} (\mathcal{L}_{T_{tr}}^{rec}(\theta)+\mathcal{L}_{T_{tr}}^{style}(\theta))$
    % \State Update $\gamma = \gamma - \beta\nabla_{\gamma}\sum_{T_j} L_{T_{j}}^{tr}(\gamma)$
    % \State return optimal $\theta^{*}$
    \EndWhile
    \State return optimal $\theta^{*}$  for  $S_{te}$ 
    \State $Style\ accuracy, bleu, domain\ accuracy = f_{T_{te}}(\theta)$
\end{algorithmic}
\end{algorithm}

%\subsection{Datasets}
\begin{table}[htbp]
\scriptsize
\centering
\begin{tabular}{@{}c|c|c|c|c|c@{}}
\toprule
Dataset    & Domain & Train & Dev & Test & Human Reference \\ \midrule
Yelp & Restaurant & 444k & 4k  & 1k  & 1k \\ \midrule
Amazon    & Product & 554k  & 2k  & 1k & 1k  \\ \midrule
IMDB & Movie & 341k  & 2k  & 1k  & No \\ \midrule
Yahoo! & Q \& A & 5k    & 1k  & 1k & No  \\ \bottomrule
\end{tabular}
\caption{Statistics of source and target datasets(non-parallel data). The style label set is \{negative, positive\}.}
\label{Table1}
\end{table}

\begin{table*}[htbp]
\centering
\scriptsize
\begin{tabular}{@{}cccccc|ccccc@{}}
\toprule
\multicolumn{6}{c|}{Restaurant(1\% target domain data)}                                                                                                                           & \multicolumn{5}{c}{Restaurant(100\% target domain data)}                                                                       \\ \midrule
Model/Training method                                     & S-Acc                & BLEU                 & G-score              & Human                & \multicolumn{1}{c|}{D-Acc}  & S-Acc                & BLEU                 & G-score              & Human                & D-Acc                \\ \midrule
CrossAlign                                & 78.4                 & 4.5                  & 18.7                    & 14.6                 & 76.8                       & 88.3                 & 5.6                  & 22.2                    & 70.3                 & 83.5                 \\
ControlGen                               & 80.1                 & 6.7                  & 23.2                    & 15.4                 & 80.4                       & 90.6                 & 25.5                 & 22.5                    & 78.9                 & 87.9                 \\
FGIM                                      & 83.1                 & 4.6                  & 19.6                    & 16.4                 & 82.0                       & 90.4                 & 24.6                 & 48.6                    & 69.4                    & 85.2                 \\

DAST                                      & 88.3                 & 17.5                 & 39.3                    & 19.5               & \textbf{90.5}                       & 91.2                & 26.5                 & 49.2                    & 79.4                    & 92.6                 \\
CatGen                                      & 85.4                 & 18.5                  & 39.7                    & 29.4                 & 80.5                       & 88.4                 & 27.9                 & 49.7                    & 65.7                   & 86.0                 \\
\textbf{ATM(ours)}                                     & \textbf{89.6}                 & \textbf{20.1}                  & \textbf{42.4}                 & \textbf{30.1}                       & 89.2                 & \textbf{93.3}                 & \textbf{30.3}                    & \textbf{53.2}                    & \textbf{85.2}       & \textbf{93.4}          \\
\hline
In-Domain                                 & 87.4                 & 9.7                  & 29.1                    & 16.4                    & 87.3                       & 94.5                 & 20.4                 & 43.9                    & 78.4                 & 93.6                 \\
Joint-Training                            & 82.3                 & 8.4                  & 26.2                    & 18.7                    & 84.6                       & 85.4                 & 21.6                 & 42.9                    & 73.6                    & 93.4                 \\
Fine-Tuning                               & 65.2                 & 2.8                  & 13.5                    & 12.6                    & 79.8                       & 92.8                 & 24.2                 & 47.3                    & 73.7                 & 93.7                 \\
D-Shift                               & 79.3                 & 10.4                  & 28.7                  &  15.4                   & 79.8                       & 91.2                & 23.4                 & 46.1                    & 73.7                 & 93.7                 \\
MAML                               & 88.2                 & 18.6                  & 40.5                    & 24.8                    & 74.5                       & 90.4               & 20.1                 & 42.6                    & 70.4                 & 92.1                 \\
\textbf{DAML(ours)}                               & \textbf{90.0}                 & \textbf{21.4}                 & \textbf{43.8}                    & \textbf{25.1}                 & \textbf{89.9}                       & \textbf{96.7}                 & \textbf{32.1}                 & \textbf{55.7}                  & \textbf{80.2}                 & \textbf{94.7}                 \\ \midrule
\textbf{DAML-ATM(ours)}                               & \textbf{94.5}                 & \textbf{25.4}                 & \textbf{48.9}                    & \textbf{34.2}                 & \textbf{92.9}                       & \textbf{97.8}                 & \textbf{35.5}                 & \textbf{58.9}                  & \textbf{83.1}                 &  \textbf{96.4}
\\
% \multicolumn{1}{l}{TSTML-zero-shot(ours)} & \multicolumn{1}{l}{} & \multicolumn{1}{l}{} & \multicolumn{1}{l}{} & \multicolumn{1}{l}{} & \multicolumn{1}{l}{}       & \multicolumn{1}{l}{} & \multicolumn{1}{l}{} & \multicolumn{1}{l}{} & \multicolumn{1}{l}{} & \multicolumn{1}{l}{} \\
\bottomrule
\end{tabular}
\caption{Evaluation results on restaurant domain(Yelp). The restaurant domain is used as the target domain and the other three domains as the source domain. G-score is the geometric mean of S-Acc and BLEU.}
\label{Table2}
\end{table*}

\begin{table*}[htbp]
\scriptsize
\centering
\begin{tabular}{@{}ccc@{}}
\hline
\textbf{}                  & Yelp(negative-to-positive)                                & Yelp(positive-to-negative)            \\ \hline
Input             & there chips are ok , but their salsa is really bland.     & love their food and their passion.    \\
Joint-Training        & there \textcolor{green}{are} good , but their \textcolor{blue}{food} is really \textcolor{red}{good}\textcolor{green}{,} . & laughable their food and \textcolor{red}{bad} \textcolor{blue}{food}.   \\
Fine-Tuning       & there chips  \textcolor{violet}{act very well}. & their food \textcolor{violet}{is hard to use}. \\
D-Shift       & there are \textcolor{blue}{usually \textcolor{violet}{dramatic} exhibits}. & \textcolor{blue}{my husband and toilet smelled}. \\
MAML              & there chips are \textcolor{blue}{bad},but there salsa is really \textcolor{red}{good}.  & \textcolor{red}{hate} their food and their passion \\
\textbf{DAML-ATM(ours)} &there chips are \textcolor{red}{surprised}, \textcolor{red}{and} their salsa is really \textcolor{red}{nice}. & \textcolor{red}{hard to swallow } food and \textcolor{red}{serious discrespect}. \\ \hline
\end{tabular}
\caption{Transferred sentences on Yelp(few shot),  where \textcolor{red}{red} denotes successful style
transfers, \textcolor{blue}{blue} denotes content losses, \textcolor{violet}{violet} denotes domain errors and \textcolor{green}{green} denotes grammar errors, better looked in color. More examples are in the appendix.}
\label{table3}
\end{table*}

\section{Experiment}
In this section, we first detail the experimental setups. Then, we present our experimental results over multiple target domains.

\subsection{Datasets and Experimental Setups}

In this experiment, we use the following four datasets from different domains: (\romannumeral1) IMDB movie review corpus~\cite{diao2014jointly}. (\romannumeral2) Yelp restaurant review dataset \cite{li2018delete}. (\romannumeral3) Amazon product review dataset \cite{li2018delete}. (\romannumeral4) YAHOO! Answers dataset \cite{li2019domain}, the amazon and yelp test sets each have 1k human annotations.The statistics of these corpora are summarized in Table \ref{Table1}. 

For the S2S model, we take the T5
base model~\cite{raffel2019exploring} (220MB) for our experiments.
For style discriminator, we use 4-layer fully connected neural networks.
We train our framework using the Adam optimizer~\cite{kingma2014adam}with the initial learning rate 1e-5. 
The epoch is set to 50 for both stage 1 and stage 2. 
The inner learning rate $\alpha$ is 0.0001, and the outer learning rate $\beta$ is 0.001. 
%Detailed configurations are provided in the supplements.
Following~\cite{shankar2018generalizing,li2020metaner}, we use the leave-one-out evaluation method by picking a domain as the target domain $D_{new}$ for the final evaluation.
For each iteration of the training phase, two source domains are randomly selected as the meta-training domain $D_{tr}$ and the remaining domains as the meta-validation domain $D_{val}$.

In order to evaluate the model performance, we use three famous and widely adopted automatic metrics following previous work~\cite{li2019domain,fu2017style,hu2017controlled}  and a human metric. \textbf{BLEU} verifies whether the generated sentences retain the original content ~\cite{papineni2002bleu}. While IMDB and Amazon have no manual references, we compute the BLEU scores w.r.t the input sentences. \textbf{Style Control} (S-Acc)  measures the style accuracy of the transferred sentences with a style classifier that is pre-trained on the datasets. \textbf{Domain Control} (D-Acc)  verifies whether the generated sentences have the characteristics of the target domain with a pre-trained domain classifier to measure the percentage of generated sentences belonging to the target domain. \textbf{Human Evaluation} Following~\cite{madotto2019personalizing}, We randomly sampled 100 sentences generated on the target domain and distributed a questionnaire at Amazon Mechanical Turk asking each worker to rank the content retention (0 to 5), style transfer(0 to 5 ) and fluency(0 to 5): $human\ score = Average(\sum score_{style}+\sum score_{content}+\sum score_{fluency}), human\ score \in [0,100]$ . 
Five workers were recruited for human evaluation. The results of the other metrics are shown in the appendix.

\subsection{Baselines}
\begin{table}[h]
\scriptsize
\centering
\begin{tabular}{@{}c|cccccc@{}}
\toprule
\textbf{Movie}   & In-Domain  & Fine-Tuning & D-Shift & MAML& \textbf{DAML} \\ \midrule
S-Acc            & 70.4      & 59.3        & 74.4 & 79.8  & \textbf{81.5}  \\ \midrule
BLEU             & 23.1      & 25.4        & 27.4 & 26.9& \textbf{31.2}  \\ \midrule
D-Acc            & 87.3      & 75.2        & 72.2 & 74.5 & \textbf{92.3}  \\ \midrule
\textbf{Product} & In-Domain & Fine-Tuning & D-Shift & MAML & \textbf{DAML} \\ \midrule
S-Acc            & 84.1      & 80.2        & 83.5  & 84.6 & \textbf{87.0}  \\ \midrule
BLEU             & 14.0       & 14.5         & 17.8  & 18.1 & \textbf{19.9}   \\ \midrule
D-Acc            & 80.5      & 75.4        & 73.5   & 79.4 & \textbf{84.1}  \\ \midrule
\textbf{Q \& A}  & In-Domain & Fine-Tuning & D-Shift & MAML & \textbf{DAML} \\ \midrule
S-Acc            & 94.1      & 90.1        & 92.1 & 89.6 & \textbf{95.5}  \\ \midrule
BLEU             & 12.8      & 13.7         & 14.5 & 18.7 & \textbf{20.5}  \\ \midrule
D-Acc            & 80.6      & 70.0        & 72.5 & 76.5 & \textbf{86.7}  \\ \bottomrule
\end{tabular}
\caption{ Results on  each of the remaining domains treated as target domain,every target domains using 1\% data for fine-tuning, base model is AMT.}
\label{Table4}
\end{table}

In our experiments, for ATM model, we adopt five state-of-the-art TST models for comparison: CrossAlign~\cite{shen2017style}, ControlGen~\cite{hu2017controlled}, DAST~\cite{li2019domain}, CatGen~\cite{wang2020cat} and FGIM~\cite{2019Controllable}. They are jointly trained on the source domains and fine-tuned on the target domain.

To well analyze our training method DAML, following~\cite{li2020metaner}, we also use five simple and effective domain adaptation settings with ControlGen~\cite{hu2017controlled} structure as DAML:
(1) \textbf{In-Domain} method is trained on the training set of the target domain; 
(2) \textbf{Joint-Training}  method combines all the training sets of the source and target domains and performs a joint-training on these datasets;
(3) \textbf{Fine-Tuning}  method is trained on the training sets of the source domains and then fine-tuned on the training set of the target domain;
(4) \textbf{D-Shift} This is trained on the combination of training sets
from all source domains. Then, the evaluation is conducted on the test set of a target domain using the direct domain shift strategy;
(5) \textbf{MAML}  method uses classical model agnostic meta-learning algorithm~~\cite{finn2017model}.

%whose model structures are similar to TSTML.
%as baselines. The model structure we use is the same as our base model of TSTML:
%\begin{itemize}
%\item \textbf{In-domain} This method is trained on the training set of the target domain. 
%\item \textbf{Joint-training} This method combines all the training sets of the source and target domains and performs a joint-training on these datasets.

%\item \textbf{Fine-tuning} This method is trained on the training sets of the source domains and then fine-tuned on the training set of the target domain.
%\end{itemize}

%In addition, we make a fair comparison with another three state-of-the-art TST models: Cross Align\cite{shen2017style}, Control-Gen\cite{hu2017controlled} and FGIM\cite{2019Controllable}.

%\subsection{Experimental Details and Hyperparameters}

%The decoder and encoder are set as single-layer Bidirectional LSTM\cite{hochreiter1997long} with hidden dimension of 300 and max sample length of 15. Discriminators are set as TextCNN\cite{kim2014convolutional}. Detailed configurations are in the supplements.

% The batch size of standard training is 32, and for meta training, the batch size was lower to 16 (5 support samples and 11 query samples). In all the models, we used beam search with beam size 1, which can be considered as equivalent to greedy search.

\begin{figure}[h]
\centering
\includegraphics[scale=0.32]{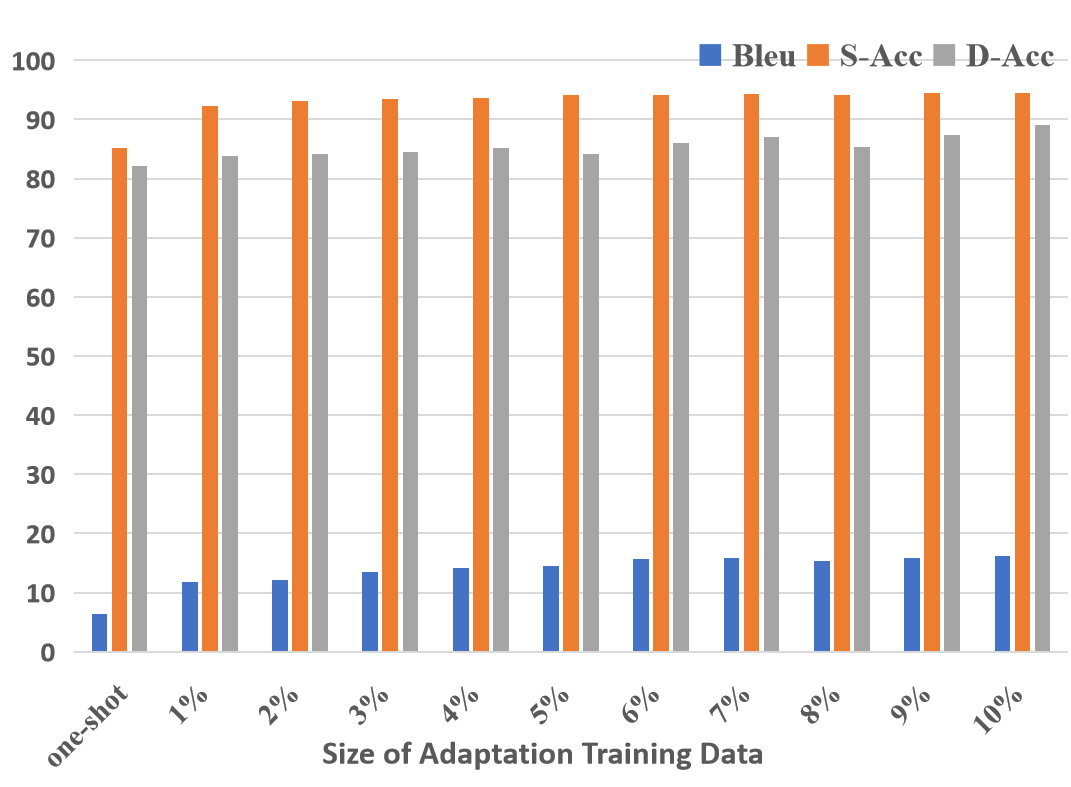}
\caption{The system performance on amazon improves when the size of the target data increases. Even the one-shot learning  achieves decent performance.}
\label{Figure4}
\end{figure}

\begin{figure}[h]
\centering
\includegraphics[scale=0.4]{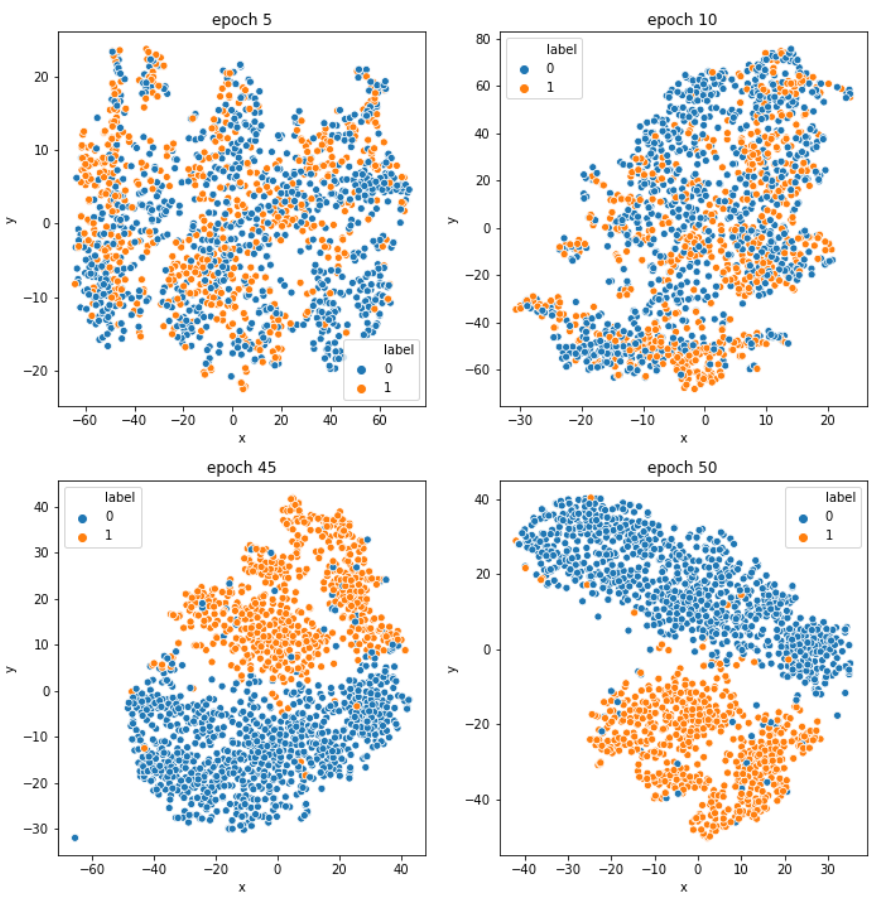}
\caption{The t-sne plots of source domain sentences and generated target domain sentence in different DAML training epochs. The labels 0 and 1 represent the source domain sentence embedding and the generated target domain sentence embedding.}
\label{Figure5}
\end{figure}

\subsection{Results and Analysis}

% % Please add the following required packages to your document preamble:
% % \usepackage{booktabs}
% % Please add the following required packages to your document preamble:
% % \usepackage{booktabs}

For DAML-ATM, we first choose restaurant as the target domain and the other three as the source domains for observation.
Table \ref{Table2} reports the results of different methods and models under both the full-data and few-shot settings. 
From this table, we can see that DAML-ATM outperforms all baselines in terms of \textit{S-Acc}, \textit{BLEU}, \textit{D-Acc} and human evaluation. We attribute this to the fact that DAML-ATM explicitly simulates the domain shift during training via DAML, which helps adapt to the new target domain.
We can also see that in the case of a few-shot setting, the results of \textit{Fine-tuning} and \textit{Joint training} are even worse than \textit{In-domain} and DAML. 
The reason may be that the data size of the source domain is much larger than the target domain so that the model tends to remember the characteristics of the source domain.
MAML achieves good performance in most metrics. However, it does not balance meta-tasks across different source domains, performing poorly on D-acc.
%, making it difficult for the model to quickly migrate to the target domain.

%  In the case that transfer learning performs so poorly on few-shot data, we achieve the highest accuracy and human score, which indicates via DATSML, the model learns some style commonality and general representation of content in different domains sentences, which enables the model to quickly adapt to the target domain. 

% Moreover, in the case of 
% few shot, the performance of the model will be reduced by using transfer learning. Because when the target domain data is scarce, using transfer learning will make the model too biased to the source domain data, which makes finetune on the target domain very difficult. Table 3 
% intuitively shows by using the TSTML across-domain parallel gradient descent, ControlGen understands the potential representations of common text styles, thus reducing grammatical errors, improving style accuracy and generating smoother sentences.

Further, to verify the robustness of our method under the low-resource setting, we separately select the other three domains as the target domain. 
%three domains of \textit{Movie}, \textit{Product} and \textit{Q\&A} 
% and use only 1\% target domain data for fine-tuning. 
As shown in Table \ref{Table4}, our approach has achieved good performance on different target domains. 
% Impressively,DAML-ST5 outperforms \textit{In-domain} by an average improvement of 5.13\%, 4.9\%, 4.9\% on \textit{S-Acc}, \textit{BLUE}, \textit{D-Acc} respectively.

%Among all combinations, TSTML outperforms the baselines in \textit{S-Acc}, BLEU and \textit{D-Acc}.
We also provide some examples in Table~\ref{table3}. 
From the example, we can see intuitively that \textit{D-shift} and \textit{Fine-tuning} will lead to the misuse of domain-specific words due to lack of target domain information. 
In addition, compared with \textit{Joint-training}, the sentences generated by DAML-ATM are more consistent with the human reference. 
Compared to MAML, DAML generates sentences that are more diverse and vivid due to the more balanced absorption of information from multiple domains.
Figure~\ref{Figure4} shows the system
performance positively correlates with the amount of training data available in the target domain. 
To visualize how well DAML-ATM performs on the new unseen domain, we use t-SNE~\cite{van2008visualizing} plots
to analyze the degree of separation between the source domain sentences and the generated target domain sentences. 
Figure~\ref{Figure5} shows that as the training epoch increases, the sentences generated by  DAML-ATM in the target domain are completely separated from the source domain in the latent space.

\subsection{Ablation Study}
To study the impact of different components on the overall performance, we further did an ablation study for our model, and the results are shown in Table~\ref{Table5}.
 After we disabled the reconstruction loss, our model failed to learn meaningful outputs and only learned to generate a word for any combination of input sentences and styles. 
Then, when the discriminator loss is not used, the model degrades rapidly, simply copying the original sentence without any style modification. 
After not using the pre-training language model weights, the model's performance is reduced in the metric of content preservation. 
When using gumble-softmax instead of hidden states for gradient descent, the model performs poorly in style accuracy because of the instability of gumble-softmax. 
In summary, each factor plays an essential role in the DAML-ATM training stage.
%  : (1) the reconstruction loss guides the model to generate readable natural language sentence. (2) the pre-trained language models encourages the model to
% preserve the information in the source sentence. (3) the discriminator provides style supervision to help the model control the style of generated sentences.
% \usepackage{booktabs}
\begin{table}[htbp]
\scriptsize
\centering
\begin{tabular}{@{}cccl@{}}
\toprule
Model                       & S-Acc & BLEU & D-Acc \\ \midrule
DAML-ATM             & \textbf{94.5}  & \textbf{25.4} & \textbf{92.9}  \\ \midrule
w/o reconstruction loss & 50.0  & 0    & 50.0  \\
w/o discriminator loss  & 2.1   & 21.6 & 92.4  \\
w/o language model weights  & 87.4   & 17.3 & 90.3  \\
w/ gumble-softmax  & 85.6   & 18.3 & 91.0  \\

% \midrule
% without DAML                & 91.2  & 25.4 & 72.3 \\ 
\bottomrule
\end{tabular}
\caption{Model ablation study results on Yelp dataset. The size of adaptation training data is 1 $\%$.}
\label{Table5}
\end{table}

% \usepackage[normalem]{ulem}

% \usepackage{booktabs}
% \usepackage{booktabs}
% \begin{table}[htbp]
% \centering
% \footnotesize
% \begin{tabular}{@{}c|c@{}}
% \toprule
% \centering
% \footnotesize
% \textbf{}            & Yelp(negative to positive)   \\ \midrule
% Input                & the food here is gross       \\ \midrule
% Human                & the food here is delicous    \\ \midrule
% Joint-training       & the food here is good to eat \\ \midrule
% Fine-tuning          & the food here is dramatic    \\ \midrule
% \textbf{TSTML(ours)} & the food here is delicous    \\ \bottomrule
% \end{tabular}
% \caption{Transferred sentences on Yelp(1\% data.)}
% \label{Table4}
% \end{table}

\section{Conclusion}
In this paper, we propose DAML-ATM, a novel training strategy combined with a new TST model for domain adaptation, which can 
be easily adapted to new domains with few shot data. 
On four popular TST benchmarks, we found significant improvements against multiple baselines, verifying the effectiveness of our method. 
We explore extending this approach for other low resource NLP tasks in future work.

\section * {Acknowledgements}
This work was partially supported by National Key Research and Development Project (2019YFB1704002) and National Natural Science Foundation of China (61876009).

\bibliography{anthology,custom}
\bibliographystyle{acl_natbib}

\appendix

\section{Appendix}
\label{sec:appendix}

\subsection{More Details on Experiment Setups}
Our model is initialized from T5 and Bart~\cite{liu2020multilingual,raffel2019exploring}. Specifically, the encoder and decoder are all 12-layer
transformers with 16 attention heads, hidden size 1,024 and feed-forward filter size 4,096, which amounts to 406M trainable parameters. We train our framework using the Adam optimizer~\cite{kingma2014adam}with the initial learning rate 1e-5, and we employ a linear schedule for the learning rate, all models are trained on 8 RTX 3090 GPUs.

\subsection{Details on Human Evaluation}
\begin{figure*}[htbp]
\centering
\includegraphics[scale=0.5]{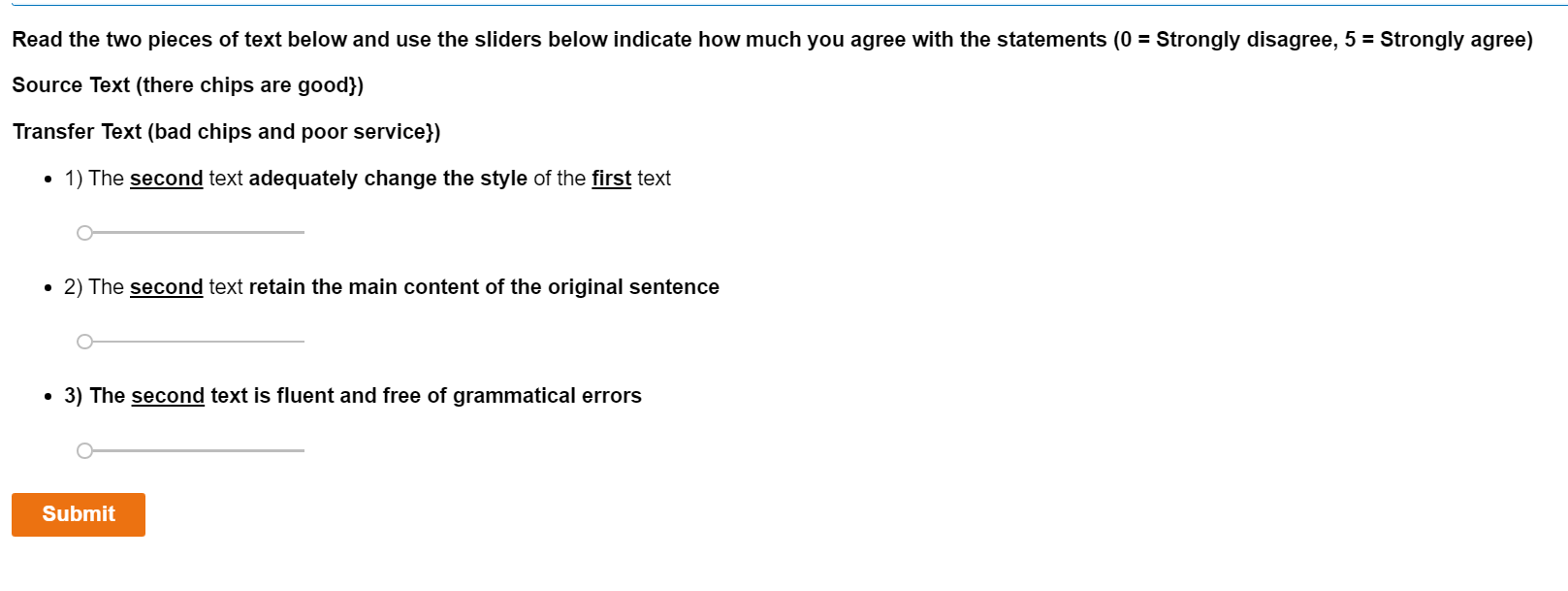}
\caption{Human evaluation questionnaire. We randomly sampled 100 sentences generated on the target domain and distributed a questionnaire at Amazon Mechanical Turk asking each worker to rank the content retention (0 to 5), style transfer(0 to 5 ), and fluency(0 to 5).}
\label{Figure6}
\end{figure*}

For the results generated by each method, following~\cite{krishna2020reformulating}, we randomly selected 100 sentences to be placed in the Amazon Mechanical Turk\footnote{https://www.mturk.com/} questionnaire. We pay our workers 5 cents per sentence. As shown in Figure~\ref{Figure6}, the questionnaire asked to judge the generated sentences on three dimensions: strength of style transfer, degree of content retention, and text fluency. To minimize the impact of spamming, we require each worker to be a native English speaker with a 95\% or higher approval rate and a minimum of 1,000 hits.

\subsection{More Ablation Study and Metrics }
To verify that the general S2S models work well with our algorithm, we use bart~\cite{lewis2019bart} as the S2S  base model. For the robustness of the experiment, we add a new metric J-(a,c,f)~\cite{krishna2020reformulating} to measure our results, which is a sentence-level aggregation strategy evaluate style transfer models.
% \usepackage{booktabs}
\begin{table}[htbp]
\scriptsize
\centering
\begin{tabular}{@{}llllll@{}}
\toprule
Domain                & S-Acc & BLEU & G-score & D-Acc & J-(a,c,f) \\ \midrule
Restaurant(T5-base)   & 94.5  & 25.4 & 48.9    & 89.2  & 46.4      \\
Restaurant(Bart-base) & 94.7  & 24.1 & 47.8    & 88.4  & 40.2      \\
Movie(T5-base)        & 81.5  & 31.2 & 50.4    & 92.3  & 42.8      \\
Movie(Bart-base)      & 84.5  & 34.7 & 54.1    & 90.1  & 43.5      \\
Product(T5-base)      & 87.0  & 19.9 & 41.6    & 84.1  & 34.5      \\
Product(Bart-base)    & 84.3  & 20.4 & 41.4    & 86.4  & 34.7      \\
Q \& A(T5-base)       & 95.5  & 20.5 & 44.25   & 86.7  & 39.5      \\
Q \& A(Bart-Base)     & 92.5  & 17.7 & 40.46   & 79.8  & 34.1      \\ \bottomrule
\end{tabular}
\caption{Results on each of the remaining domains treated as target domain, every target domain using 1\% data for fine-tuning, base models are BART and T5.}
\label{Table6}
\end{table}

As can be seen from Table 6, our approach can be combined with other general pre-trained language models and performs well, proving our method's generality. 
Furthermore, as we can visually see from Table 7, our model also performs well on the J-(a,c,f) metric, which indicates that our model generates sentences in a specific style while having the right target style, preserving content, and being fluent.

\subsection{More Generation Examples}
To demonstrate more examples of generation to verify the effectiveness of the model, we selected 10 generated sentences from amazon and yelp each, as shown in Table~\ref{Table9} and Table~\ref{Table10}.

\begin{table}[H]
\scriptsize
\centering
\begin{tabular}{@{}ccc@{}}
\toprule
\multicolumn{2}{c|}{Restaurant(1\% target domain data)}                     & Restaurant(100\% target domain data) \\ \midrule
\multicolumn{1}{c|}{Model/Training Method} & \multicolumn{1}{c|}{J-(a,c,f)} & J-(a,c,f)                            \\ \midrule
 CrossAlign                                          & 18.4                           & 22.9                                 \\
                  ControlGen                         & 19.2                           & 24.5                                 \\
                   FGIM                        & 25.6                           & 28.7                                 \\
                     DAST                      & 24.5                           & 32.3                                 \\
                  Cat-Gen                         & 20.3                           & 31.2                                 \\
\textbf{ATM(ours)}                                  & \textbf{30.4}                           & \textbf{39.5}                                 \\ \midrule
             In-Domain                              & 32.5                           & 35.2                                \\Joint-Training
                                           &   32.3                             &  35.4                                    \\ Fine-Tuning
                                           &    28.2                            &  37.8                                    \\D-Shift
                                           &    25.5                        &      34.1                                \\MAML
                                           &    38.8                            &   44.1                                   \\\textbf{DAML(ours)}
                                           &     \textbf{44.8}                           &   \textbf{47.6}                                   \\ \midrule
\textbf{DAML-ATM(ours)}                             &   \textbf{46.4}                             &   \textbf{50.8}                                    \\ \bottomrule
\end{tabular}
\caption{Evaluation results on restaurant domain(Yelp). The restaurant domain is used as the target domain and the other three domains as the source domain. G-score is the geometric mean of S-Acc and BLEU.The evaluation metrics is J-(a,c,f)}
\label{Table7}
\end{table}

\begin{table*}[htbp]
\centering
\footnotesize
\begin{tabular}{l|l}
\hline
          & Amazon (negative-to-positive) or   (positive-to-negative)                                                                               \\ \hline
Input      & anker, don't like it.                                \\
DAML-ATM     & anker is a great product, i like it.                                       \\
Human Ref. & anker is a fine product though i d never heard of it.                                     \\ \hline
Input      & but the back side looks terrible as well. \\
DAML-ATM     & but the back side looks great as well.                                    \\
Human Ref. & but the back side looks nice as well .                                   \\ \hline
Input      & the hot dogs come out bad                                     \\
DAML-ATM    & the hot dogs come out great.                                                      \\
Human Ref. & we often cook sausages ,  hot dogs and hamburgers in it as well .                                                   \\ \hline
Input      & i will buy hamilton beach products again .                                                  \\
DAML-ATM     & i will not buy hamilton beach products again.                                               \\
Human Ref. & i will never buy hamilton beach products again .                                            \\ \hline
Input      & synching is not too hard , with no passwords or codes required                     \\
DAML-ATM     & synching is easy, with no passwords or codes required.                  \\
Human Ref. & synching is simple ,  with no passwords or codes required .                 \\ \hline
Input      & they were flimsy       \\
DAML-ATM     & they were a bit flimsy, but they were ok.                             \\
Human Ref. &  was concerned for the price that these measuring cups would be flimsy .                            \\ \hline
Input      & it does not move much air.                                              \\
DAML-ATM      & it moves air very fast.                                                   \\
Human Ref. & it moves plenty of air ,  oscillating between two rooms .                                                  \\ \hline
Input      & very ugly unit, hard to assemble and difficult to operate.                  \\
DAML-ATM     & very nice unit, easy to assemble and operate.                         \\
Human Ref. & very nice unit ,  easy to assemble and operate .                       \\ \hline
Input      & i would recommend this money case                                                           \\
DAML-ATM      & i would not recommend this money case to anyone.                                            \\
Human Ref. & i would never recommend this terrible waste of money case.                                  \\ \hline
Input      & And this is not what you would need.                                  \\
DAML-ATM      & and this is what you would need.                                              \\
Human Ref. & and that would be what you would need .                                            \\ \hline

\end{tabular}
\caption{More Transferred sentences on Amazon(few shot setting).}
\label{Table9}
\end{table*}

\begin{table*}[htbp]
\centering
\footnotesize
\begin{tabular}{l|l}
\hline
          & Yelp  (negative-to-positive) or   (positive-to-negative)                                                                   \\ \hline
Input      & Ever since joes has changed hands it's gotten better and better.         \\
DAML-ATM     & ever since joes has changed hands it's gotten worse and worse.           \\
Human Ref. & ever since joes has changed hands it 's just gotten worse and worse .    \\ \hline
Input      & There is so much room in that part of the venue                          \\
DAML-ATM     & there is not enough room in that part of the venue.                      \\
Human Ref. & there is definitely not enough room in that part of the venue .          \\ \hline
Input      & It didn't taste watered down at all.                                     \\
DAML-ATM      & it tasted watered down.                                                  \\
Human Ref. & so basically tasted watered down .                                       \\ \hline
Input      & This pharmacy is really considerate.                                     \\
DAML-ATM      & this pharmacy is really a shithole.                                      \\
Human Ref. & i ca n't believe how inconsiderate this pharmacy is .                    \\ \hline
Input      & definitely not disappointed that i could use my birthday gift !          \\
DAML-ATM      & definitely disappointed that i could not use my birthday gift!           \\
Human Ref. & definitely disappointed that i could not use my birthday gift !          \\ \hline
Input      & but it probably doesn't suck too !                                       \\
DAML-ATM      & but it probably does suck too!                                           \\
Human Ref. & but it probably sucks too !                                              \\ \hline
Input      & the service was quick and responsive                                     \\
DAML-ATM      & the service was slow and not responsive.                                 \\
Human Ref. & we sit down and we got some really slow and lazy service .               \\ \hline
Input      & they said we could sit at the table with no hesitation                   \\
DAML-ATM      & they said we could not sit at the table.                                 \\
Human Ref. & said we could n't sit at the table if we were n't ordering dinner .      \\ \hline
Input      & the wine was above average and the food was even better                  \\
DAML-ATM     & the wine was average and the food was even wore.                        \\
Human Ref. & the wine was very average and the food was even less .                   \\ \hline
Input      & i would not visit this place again           \\
DAML-ATM      & i would definitely visit this place again.   \\
Human Ref. & one of my favorite chinese place to eat ! \\ \hline
\end{tabular}
\caption{More Transferred sentences on Yelp(few shot setting).}
\label{Table10}
\end{table*}

\end{document}